\pdfoutput=1

\documentclass[11pt,a4paper]{article}
\usepackage[hyperref]{eacl2021}
\usepackage{times}
\usepackage{latexsym}

\usepackage{microtype}

\usepackage{url}
\usepackage{relsize}
\usepackage{hyperref}
\usepackage{tabu}
\usepackage{xspace,amsmath,amsfonts}
\usepackage{multirow,multicol}
\usepackage{mathtools}

\usepackage{tikz} 
\usepackage{pgfplots}
\usepackage{subcaption}
\usepackage{tabularx}

\usetikzlibrary{pgfplots.groupplots}

\usepackage{enumitem}
\setitemize{noitemsep,topsep=0pt,parsep=0pt,partopsep=0pt}

\usepgfplotslibrary{groupplots}

\aclfinalcopy % Uncomment this line for the final submission
\def\aclpaperid{970} %  Enter the acl Paper ID here

\DeclarePairedDelimiter{\norm}{\lVert}{\rVert} 

\newcommand{\relic}{\textsc{RELIC}\xspace}
\newcommand{\taggenome}{\textsc{TagGenome}\xspace}

\newcommand{\bert}{\textsc{BERT}\xspace}
\newcommand{\berty}{\textsc{DocEnt}\xspace}
\newcommand{\bertysimple}{\textsc{DocEnt-Full}\xspace}
\newcommand{\bertysimpleabbr}{\textsc{Full}\xspace}
\newcommand{\bertydual}{\textsc{DocEnt-Dual}\xspace}
\newcommand{\bertyhybrid}{\textsc{DocEnt-Hybrid}\xspace}

\newcommand{\sbert}{\textsc{SentenceBERT}\xspace}
\newcommand{\bagofglove}{\textsc{BoS-GloVe}\xspace}
\newcommand{\bagofbert}{\textsc{BoS-BERT}\xspace}
\newcommand{\bagofbertave}{$\textsc{BoS-BERT}^*$\xspace}
\newcommand{\bagofsbert}{\textsc{BoS-SentenceBert}\xspace}
\newcommand{\tfidf}{\textsc{TF-IDF}\xspace}
\newcommand{\toptags}{MovielensTopTags}
\title{\berty: Learning Self-Supervised Entity Representations from Large Document Collections}

\author{Yury Zemlyanskiy\Thanks{ Work is partially done while at Google} \\
  U. of Southern California \\
  \small{\texttt{yury.zemlyanskiy@usc.edu}} \\
  \And
  Sudeep Gandhe \\
  Google Research \\
  \small{\texttt{srgandhe@google.com}} \\
  \And
  Ruining He \\
  Google Research \\
  \small{\texttt{ruininghe@google.com}} \\
  \And
  Bhargav Kanagal \\
  Google Research \\
  \small{\texttt{bhargav@google.com}} \\
  \AND
  Anirudh Ravula \\
  Google Research \\
  \small{\texttt{braineater@google.com}} \\
  \And
  Juraj Gottweis \\
  Google Research \\
  \small{\texttt{juro@google.com}} \\
  \And
  Fei Sha\Thanks{ On leave from USC (feisha@usc.edu)}    \\
  Google Research \\
  \small{\texttt{fsha@google.com}} \\
  \And
  Ilya Eckstein \\
  Google Research \\
  \small{\texttt{ilyaeck@google.com}}}

\date{}
\pgfplotsset{
  compat=1.15,
  every axis plot/.append style={line width=1pt},
  every axis plot post/.append style={
    every mark/.append style={line width=1.6pt,draw=green,fill=red}
  }
}

\begin{document}
\maketitle
\begin{abstract}
This paper explores learning rich self-supervised entity representations from large amounts of associated text. Once pre-trained, these models become applicable to multiple entity-centric tasks such as ranked retrieval, knowledge base completion, question answering, and more. Unlike other methods that harvest self-supervision signals based merely on a local context within a sentence, we radically expand the notion of context to include {\em any} available text related to an entity. This enables a new class of powerful, high-capacity representations that can ultimately distill much of the useful information about an entity from multiple text sources, without any human supervision. 

We present several training strategies that, unlike prior approaches, learn to {\em jointly} predict words and entities---strategies we compare experimentally on downstream tasks in the TV-Movies domain, such as MovieLens tag prediction from user reviews and natural language movie search. As evidenced by results, our models match or outperform competitive baselines, sometimes with little or no fine-tuning, and can scale to very large corpora.

Finally, we make our datasets and pre-trained models publicly available\footnote{See \url{http://goo.gle/research-docent} for {\em Reviews2Movielens} and models. Scripts and {\em Reddit Suggestions} can be found at \url{https://urikz.github.io/docent}}. This includes {\em Reviews2Movielens}, mapping the {\raise.17ex\hbox{$\scriptstyle\mathtt{\sim}$}}1B word corpus of Amazon movie reviews \cite{amazonreviews} to MovieLens tags~\cite{harper2016movielens}, as well as Reddit Movie Suggestions with natural language queries and corresponding community recommendations.
\end{abstract}

% !TEX root = main.tex
\section{Introduction}

\begin{table}[t!]
\centering
\begin{tabular}{ | m{0.96\columnwidth} | }
\hline
 {\bf \small Review 1: }{\it \small
``This movie develops its power best if you don't try to look out for the ``real'' and ``true'' events behind the four versions of the narration... shown in a very intelligent and artistic way, no silly plot-twists, no explanation in the end --- it is open to your fantasy... ``\textsc{\$movie}'' is an important piece of cinematic storytelling and a really interesting way to reflect on the origin of tales... Some scenes even remind me of Andrej Tarkovskijs intensive style..''.} \\
\hline
  {\bf \small Review 2: }{\it \small
``Just rented this, and at first I didn't like very much, but then it starts to sink in for how good it is, the acting is great especially Toshiro Mifune, it was shot very good for an older movie... it’s \#62 on the top 250''} \\
\hline
  {\bf \small Review 3: }{\it \small
``Saw this movie at my local video store... was placed on a waiting list, but when I returned to check it out the video store had closed down over night. Actually whent out of business''} \\
\hline
{\small \hspace{2.5cm} ... More reviews ...} \\
\hline
{\small {\bf Summary tags:} [nonlinear] [multiple storylines] [japan] [black and white] [surreal] [cerebral] [imdb top 250], ...} \\
\hline
\end{tabular}
\caption{\label{table:rashomon_example} \small {\em Reviews2Movielens} task, illustrated. Here are sample review snippets for a certain classic film which is summarized using MovieLens tags. Notice that the tags may not appear in the input verbatim and can be thought of as boolean questions about the film. Note also that Review 3 has zero relevant signal---a common challenge of low SNR in this dataset. \\
Bonus teaser: can you guess the \textsc{\$movie} from these snippets? This little quiz alludes to a key learning task in our approach.}
\end{table}

Much of the online information describing entities in domains such as music, movies, venues or consumer products, is only available as unstructured text---a format that is human-readable but not machine-understandable (yet).
Consider online reviews---a rich source of mostly user-generated about a vast number of entities.
Our key research question is: \emph {Can we learn strong models for entity understanding tasks such as vertical search and question answering, solely from text?}
In other words, given a large and noisy collection of documents about an entity, can we distill all the useful information therein into a dense entity representation, so as to benefit multiple downstream tasks? 

Traditionally, learning entity representations required supervised signals such as clicks, ``likes" and consumption behavior~\citep{agichtein2006improving,huang2013learning,koren2009matrix,vig2012tag}, which are generally expensive and time consuming to obtain at scale. 
To leapfrog these limitations, we draw inspiration from the recent progress in unsupervised learning of text, particularly contextualized representations via techniques such as ELMo~\citep{elmo}, CoVe~\citep{cove} and BERT~\citep{devlin2018bert}.
Many of these representations are learned by predicting a missing word from its context.
More recently, \citet{DBLP:journals/corr/abs-1904-09223} showed that extending word masking strategies to entities can lead to superior language models. 
Even more recent entity linking methods such as \relic~\citep{relic} and others, detailed in Section~\ref{related_work}, were shown to produce explicit encodings applicable to entity understanding tasks.

We start with \relic-like approaches and generalize them into a family of models, collectively called \berty, that jointly embed text and entities (Section~\ref{method}) via self-supervised tasks.
The first one, \bertydual, is essentially \relic, but trained with a much broader context to include any and all sentences potentially related to an entity. Importantly, \bertydual/\relic only optimizes a single task, namely {\em entity prediction} given an associated sentence, effectively modeling $P(Entity | Sentence)$.

Another natural way of jointly modelling entities and text is by directly tapping the cross-attention mechanism in \bert, simply by extending the \bert vocabulary to include entity tokens $V_E$.
Each entity-related sentence can then be augmented with a corresponding token from $V_E$. 
We call this method \bertysimple and, despite (or perhaps because of) its conceptual simplicity, it proves surprisingly effective in semi-supervised tasks.

Finally, \bertyhybrid aims to capture the best of both models by extending \bertydual with an additional task of predicting words in a sentence, conditioned on its associated entity. This task encourages the latter to ``remember'' salient phrases in its sentences.

We empirically evaluate these methods by learning entity representations for movies from a TV-Movies portion of the Amazon Reviews Corpus~\citep{amazonreviews}. To this end, we consider several movie-oriented tasks for downstream evaluation, i.e. Reddit Movie Suggestions and MovieLens Tag Prediction~\citep{harper2016movielens}, which we study in both zero-shot, supervised and few-shot settings. We join the MovieLens dataset with the reviews corpus~\citep{amazonreviews} obtaining a mapping from movie reviews to user-generated tags. On the supervised tag prediction task, our text-based model demonstrates SOTA performance, despite not using powerful user signals~\citep{vig2012tag}. In fact, we are able to match or outperform baselines on all tasks where they are available.

\subsection{Contributions}
\begin{enumerate}
\item First, we propose a family of methods to train deep self-supervised entity representations purely from related text documents, with strong zero-shot results on ranked retrieval with natural language queries. 
\item Secondly, we show that these pre-trained representations are amenable to fine-tuning on new tasks such as MovieLens tag prediction, where we show state-of-the-art results. They are also effective few-shot learners, which we demonstrate on a harder \emph{open-vocabulary}\footnote{An open vocabulary allows any phrase to be a label.} task akin to Boolean Question Answering \citep{clark-etal-2019-boolq}.
\item Next, we propose {\em Reviews2Movielens}---a new Text Based Entity Understanding task. The requisite dataset, which we release publicly, effectively joins the Amazon Movie Reviews Corpus and MovieLens into a large, sparsely supervised set with approximately 1B words and 470K movie-tag pairs.
\item Finally, we also release a dataset of user-generated Reddit Movie Suggestions, a benchmark for natural language search and recommendation scenarios.  
\end{enumerate}

% !TEX root = main.tex

\section{Self-Supervised Entity Representations}
\label{method}
\begin{figure*}
    \centering
    \includegraphics[width=\textwidth,bb=0 0 2692 714]{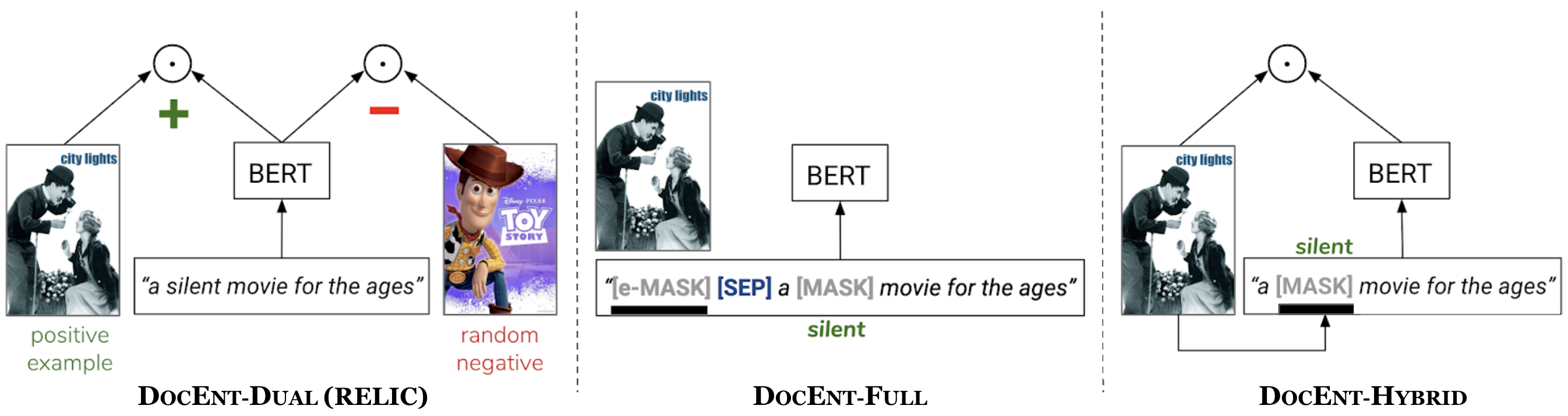} 
    \caption{\small Models in the \berty family. Left: a baseline dual encoder model called \bertydual a.k.a. \relic, maximizing $P(e | s)$ but not $P(s | e)$. Center: \bertysimple---a model maximizing the joint sentence-entity probability using full cross-attention. Right: \bertyhybrid, designed to capture the best of both worlds.}
    \label{fig:berty}
\end{figure*}

Inspired by the success of self-supervised language models, we seek to extend them to jointly compute text and entity representations.
Recall that our input is a set of entities $\mathcal{E}$ where for every entity $e \in \mathcal{E}$, we have a collection of sentences, denoted by $\mathcal{S}_e$, from all documents related to $e$. 
Intuitively, we want the representation of $e$ to be influenced by each associated sentence $s \in \mathcal{S}_e$, and vice versa. To that end, we explore two (self-) supervision signals: $P(e \mid s)$ and $P(s \mid e)$.

\subsection{\bertydual, Known as \relic}
\label{sec:bertydual}
At the core of \bertydual is a \relic model that co-encodes an entity $e$ and an associated sentence $s \in \mathcal{S}_e$ so as to maximize their compatibility score, defined as the cosine similarity between the two encodings: 
\begin{align*}
    \textbf{s}(e, s) = \frac{g(e)^Tf_{CLS}(s)}{\norm{g(e)}\norm{f_{CLS}(s)}},
\end{align*}
where $g(e)$ is an embedding of $e$ and $f(s)$ is a \bert-based encoding of s, with its special $[CLS]$ token whose output representation is denoted by $f_{CLS}$. Then, the conditional probability of $e$ given $s$ is given by a softmax over the set $\mathcal{E}$~\footnote{In practice, only a subset of entities in $\mathcal{E}$ is used in the denominator: the so called ``in-batch negatives''.}:
\begin{align*}
    P(e | s) = \frac{\exp(\textbf{s}(e, s))}{\sum_{e' \in \mathcal{E}} \exp(\textbf{s}(e', s))}.
\end{align*}
Finally, \relic is trained by maximizing $\log P(e | s)$ over all associated pairs $e, s \in \mathcal{S}_e$:
\begin{align*}
\mathcal{L}_{E}(e, s) = \log P(e | s).
\end{align*}
Note that both $g$ and $f$ (initialized with a common \bert) are learned during training. 

Our sole difference to the original \relic is in training data: while \relic only uses sentences containing entity mentions, we allow a radically broader context -- all sentences associated with an entity -- with the goal of remembering all of its attributes. Crucially, no human labeling is required.

Despite its effectiveness (as demonstrated in Section~\ref{sec:experiments}), \relic has one obvious limitation: it ignores $P(s \mid e)$, leaving a useful signal ``on the table". We therefore propose another way of co-encoding sentences and entities by tapping the full cross-attention power of Transformers. 

\subsection{\bertysimple}
\label{sec:bertysimple}
Before we proceed, let us revisit \bert's Masked Language Model (MLM) training objective. Given a sequence of input tokens $s = [s_1, \ldots, s_n]$, a fraction of tokens $s_J$ at randomly selected positions $J$ is replaced with a special \textit{[MASK]} token. We denote this new sequence by $s_{-J}$.

Then, \bert predicts masked tokens based on their contextualized representations $f(s_{-J})$. The MLM training objective to maximize is:
\begin{align*}
    \mathcal{L}_{MLM} = \log P(s_J \mid s_{-J}).
\end{align*}
Enter \bertysimple. It follows the standard \bert architecture, with a twist. 
First, we expand the input vocabulary to include all entity tokens in $\mathcal{E}$. 
Then, during input sequence construction, each sentence $s \in \mathcal{S}_e$ is prepended\footnote{Technically, we replace \bert's standard ($s_A$, $s_B$) two-segment input structure with ($e$, $s$), for $s \in \mathcal{S}_e$.} with the corresponding entity token $e$, as shown in Figure~\ref{fig:berty}. This way, masking and predicting this token (via softmax) effectively adds our new objective $\mathcal{L}_{E}$ to \bert. Further, the new $e$ token is now part of a sentence context, augmenting the original $\mathcal{L}_{MLM}$ to
\begin{align*}
    \mathcal{L}_{MLM+E}(s, e) &= \log P(s_J \mid s_{-J}, e), \\
    \text{and       } \mathcal{L_{\bertysimpleabbr}} &= \mathcal{L}_{E} + \lambda \mathcal{L}_{MLM+E}
\end{align*}
becomes the combined loss function optimized using nothing but \bert's standard MLM training, with a hyperparameter $\lambda$ to balance the two terms\footnote{The relative masking frequency of entity tokens is another hyperparameter available to balance the two objectives.}.

This conceptual simplicity and full cross-attention power come with a cost: bundling wordpieces and entities together forces the model to allocate an equal capacity to both types of tokens (e.g., 768D for \bert-base), regardless of the size of $\mathcal{E}$. As a result, a relatively small-sized $\mathcal{E}$ may be prone to overfitting\footnote{Conversely, a very large $\mathcal{E}$ may require an optimized implementation of softmax to maintain scalability.} in zero-shot scenarios, as we observe in Section~\ref{eval:reddit}.

\subsection{\bertyhybrid}
\label{sec:bertyhybrid}
Recall that \relic avoids the above limitation by decoupling text and entity encoders. To get the best of both worlds, we introduce \bertyhybrid---a third model that sticks with the modular dual encoder architecture while also modeling \textbf{$P(s \mid e)$}.
This is achieved by implementing a different variant of $\mathcal{L}_{MLM+E}$ where, for every masked wordpiece token, the output of Transformer layers $f(s_{-J})$ is first concatenated with the associated entity embedding $g(e)$ before feeding into the final MLM prediction layer.
By including entity embeddings in the prediction of related text tokens, we get them to ``remember" important aspects from the text without sacrificing modularity.

% !TEX root = main.tex
\section{Tasks}
\label{sec:tasks}

In this section, we define the three tasks used to evaluate pre-trained entity representations.

\subsection{Supervised Task: Movielens Tag Prediction}
The original MovieLens Tag Prediction task is to produce movie-tag scores for a set of movies and a canonical vocabulary of tags (see examples in Table~\ref{table:rashomon_example}), based on a collection of crowdsourced (movie, tag, user) votes, as well as (user, movie) star ratings. These tags are often not factual but may refer to plot elements, qualitative aspects or reflect subjective opinions. Since the same can be said about user reviews, and we observe a non-trivial amount of textual entailment between the two sources. We therefore intentionally exclude user ratings from the input. The new challenge is to complete the movie-tag relevance matrix by leveraging movie reviews, hereafter referred to as the {\em closed-vocabulary tag prediction} task\footnote{One can also view this as a two-dimensional knowledge base (KB) completion problem, where relation types are not available and the KB is reduced to a 2D matrix.}.
This is a supervised setup where models are fine-tuneed with tag labels and evaluated on a held-out set subset of movies, as elaborated in Section~\ref{sec:experiments}. 

\subsection{Few-Shot Task: Open Vocabulary Tag Prediction}
In reality, the space of tags is not static. Rather, tags are a useful kind of user-generated content that evolves to reflect the zeitgeist, much like human language. Many online platforms (e.g, Twitter and Instagram to name a few) have vibrant online communities that keep inventing new tags. We therefore propose a new {\em open-vocabulary} formulation of the tag prediction problem where any phrase is allowed to be a tag.

This requires a small change in evaluation. Instead of held-out movies, we hold out a subset of tags and fine-tune on the rest (and on all the movies). Note that this is no longer a classic multi-label classification task as we never get to see the test labels during training. Rather, this open-vocabulary setup is akin to answering boolean questions (about a movie) based on a text document~\cite{clark-etal-2019-boolq}.

% !TEX root = main.tex

\newcommand{\light}[1]{\textcolor{gray}{#1}}

\begin{table*}[th]
\centering
\begin{tabularx}{\textwidth}{ 
>{\arraybackslash}p{8.6cm} |
>{\arraybackslash}p{6.5cm} }
 \textbf{Query} & \textbf{Top 5 Results} \\
 \hline
  {\small Movies like [\light{{\bf Whiplash}}] about an artist or a musician chasing an almost  impossible dream and nearly or does ruin his life because of it} & {\small Inside Llewyn Davis, {\bf Whiplash}, A Young Man with a Horn, Hustle \& Flow, Born to Be Blue } \\
 \hline 
  {\small Really dark, slow paced movies with minimal story, but incredible atmosphere, kinda like [\light{\bf Drive}] or [\light{{\bf The Rover}}}] & {\small {\bf The Rover}, Valhalla Rising, Only God Forgives, Blade Runner, Sicario } \\
  \hline
  {\small Films like [\light{{\bf Mission Impossible}}] or [\light{{\bf The Italian Job}}] that have big scenes where the characters must break in or infiltrate some place } &
 {\small National Treasure: Book of Secrets, {\bf Mission: Impossible – Rogue Nation}, Ant-Man, {\bf The Italian Job }} \\
 \hline
\end{tabularx}
\caption{\label{table:reddit_qual} \small Qualitative examples illustrating zero-shot movie ranking by \bertysimple, with natural language queries crawled from Reddit. 
The bracketed greyed-out movie mentions are users' examples of desired recommendations, removed from the queries to probe the model in what resembles a movie guessing game. 
Those obfuscated entities were correctly guessed by the model based on remaining query terms, making it to the Top 5 in most cases. Other top matches appear to be equally relevant.
}
\end{table*}

\subsection{Zero-Shot Task: Reddit Movie Suggestions}
\label{sec:tasks:reddit-zero-shot}
The purpose of this task is to evaluate pre-trained entity representations in the context of vertical search. The classic entity ranking problem is, given a text query and a finite set of entities, to rank them according to their relevance to the query. Recall that \berty models are naturally designed to make such relevance predictions via $P(Entity | Sentence)$ --- without any fine-tuning, if necessary.
We therefore leverage the Reddit Movie Suggestions Dataset (detailed in Section~\ref{datasets:reddit}) as a source of both queries and ground truth to define a zero-shot movie ranking task. 
To clarify, the notion of \emph{zero shot} implies a pre-trained but not fine-tuned model in our context.     
This dataset is particularly interesting for its challenging queries, with their distinctly natural, often conversational language (e.g., \textit{\small ``Last week I watched the British cold war movie Threads. I am scarred, but intrigued as well. Any similar deeply disturbing yet realistic movies you can recommend?''}, see Table~\ref{table:reddit_qual} for more examples). Another challenge is an explicit recommendation intent present in many of the queries (i.e., \textit{\small ``Movies like ...''}), making this task a mixture of Search and Recommendation. The latter typically requires specialized recommendation models of entity-to-entity similarity, and cannot generally be solved with keyword-based search.

% !TEX root = main.tex
\section{Datasets}
\label{datasets}

\subsection{Amazon Movie Reviews Corpus}
All our models are pretrained on Amazon Product Reviews~\citep{amazonreviews} in the ``Movies and TV'' category, comprising 4,607,047 reviews for 208,321 movies collected during 1996--2014\footnote{We've used the 2016 version of the dataset from \url{http://jmcauley.ucsd.edu/data/amazon}.}.

\subsection{Reviews2Movielens}
One of this paper's contributions is {\em Reviews2Movielens}---a new multi-document multi-label dataset created by joining Amazon Movie Reviews~\citep{amazonreviews,ni-etal-2019-justifying} and MovieLens~\citep{harper2016movielens}, a rich source of crowdsourced movie tags. The key challenge in joining the two datasets is establishing correspondences between their respective movie IDs, which turns out to be a many-to-one mapping\footnote{Each Amazon ID (ASIN) matches a canonical product URL, e.g., \url{https://www.amazon.com/dp/B06XGG4FFD}. However, these IDs correspond to specific product {\em editions} (typically DVDs) rather than unique titles, causing duplication issues.
Some are collections of several titles.}.
We have identified a subset of high-precision many-to-one correspondences by applying Named Entity Recognition techniques\footnote{We use the public Google Cloud Natural Language API -- \url{https://cloud.google.com/natural-language/docs/basics\#entity\%20analysis}.} to both Amazon product titles (incl. release years) and their product pages. 
The resulting mapping consists of 71,077 unique Amazon IDs and 28,918 unique MovieLens IDs. The mapping accuracy was manually verified to be 97\% based on 200 random samples.
Ultimately, the joined dataset contains nearly 2 million reviews and close to 1B words, significantly more than its IMDB counterpart~\citep{imdbreviews}.

Since both datasets are widely used as a source of data and academic benchmarks~\cite{miller2003movielens,jung2012attribute,anand2016semi,amazonreviews,ni-etal-2019-justifying}, we hope that this new mapping\footnote{See \url{http://goo.gle/research-docent}} will be useful to the community. 

\subsection{Reddit Movie Suggestions}
\label{datasets:reddit}
This user-generated dataset contains a collection of 4765 movie-seeking queries and corresponding recommendations, collectively curated and voted on by the Reddit Movie Suggestions community\footnote{\url{https://www.reddit.com/r/MovieSuggestions}}. Worth noting are (a) the conversational, human-to-human language of the queries; (b) the community-recommended movies that, while sparse and possibly biased, can be used as a source of ground truth. While modest in size, the dataset is well-suited to evaluate zero-shot performance on the movie ranking task defined in Section~\ref{sec:tasks:reddit-zero-shot}.

% !TEX root = main.tex
\section{Experiments}
\label{sec:experiments}

\subsection{Pre-training}
All our experiments start with pre-training models on the Amazon Movie Reviews corpus, followed by optional task-dependent fine-tuning.
First, we apply some simple filtering to the input, removing reviews shorter than 5 words and movies with less than 5 reviews~\footnote{This low-count filtering is applied after de-duplication and aggregation.}. This results in 81,057 Amazon movies, of which 17,131 have MovieLens correspondences, and 4,181,727 reviews in total. Further, we split reviews into individual sentences (or short paragraphs) so as to circumvent the \bert sequence length limit. Finally, since our goal is to learn non-obvious entity attributes, we remove movie names from their reviews.

All our models use the standard BERT-base configuration with 12 layers, 12 attention heads and a hidden size of 768, and are initialized with a publicly available BERT-base checkpoint\footnote{\url{https://storage.googleapis.com/bert_models/2018_10_18/uncased_L-12_H-768_A-12.zip}}.

\subsection{Tag Prediction: Fine-tuning Strategies}
\label{sec:tasks:tag-finetuning}
We will now describe the fine-tuning strategies used to transfer pre-trained \berty models to downstream tag prediction tasks. 

\paragraph{\bertysimple}
To generate movie-tag relevance scores, we need to predict $P(Tag | Movie)$, which we cast as binary classification. Recall that \bert has a built-in binary classifier (for next-sentence prediction), implemented as a single-layer FFN~\footnote{Feed-Forward Neural Network} on top of its \textit{[CLS]} output, with logistic loss. We simply repurpose that layer for our task.

\paragraph{\bertydual and \bertyhybrid}
Recall that, during pre-training, \bertydual and \bertyhybrid use softmax cross entropy loss to predict $P(Entity | Sentence)$. However, tag prediction poses the inverse problem: predict tags based on a movie entity. In our dual encoder framework, that can be done simply by computing softmax over all of the encoded tags rather than entities, without any changes to the architecture. 

\paragraph{Shared Strategies}
For fine-tuning, all of the models share the following choices. First, we treat every existing movie-tag pair in the training set as a positive example, weighted proportionally to the number of user votes for that pair (or to the logarithm thereof). Next, for a given movie, about $~10\%$ of all vocabulary tags are sampled as negative examples, excluding the known true positives for that movie. 
To prevent overfitting, we fix entity embedding weights for all models during fine-tuning.

\subsection{Entity-less Baselines}
\label{eval:baselines}
To corroborate the utility of explicit entity representations, we set out to evaluate a few baselines that circumvent them by representing each entity as a Bag-of-Sentences (BoS), computed over its related reviews with a sentence encoder of choice. Such a BoS encoder can replace entity embeddings in our architecture, yielding a na\"ive variant of \bertydual. 
We call these baselines \bagofglove, \bagofbert and \bagofsbert\footnote{\sbert~\cite{sentencebert} fine-tunes \bert on NLI to provide off-the-shelf semantic sentence representations.}, reflecting their underlying sentence encoders.

\subsection{Evaluation} 
\subsubsection{Movielens Tag Prediction}
\label{eval:movielens}
% !TEX root = main.tex
\begin{table}[tp]{
\centering
\begin{tabular}{l|rrr}
\textbf{Task} & \textbf{Movies} & \textbf{Tags} & \textbf{M-T Pairs} \\
\hline
Closed (test) & 1000 & 1128 & 46359 \\
Closed (dev) & 380 & 1128 & 17943 \\
\hline
Open (test) & 6392 & 500 & 141618 \\
Open (dev) & 3362 & 100 & 25274 \\
\end{tabular}
\caption{\label{table:movielens_eval_data} \small Evaluation datasets sizes for Tag Prediction tasks. Closed / Open stand for the closed and open vocabulary tasks, respectively; M-T Pairs shows the number of corresponding movie-tags pairs. The top two rows describe \emph{movie} holdout sets used in our closed vocabulary experiments; bottom two rows showing \emph{tag} holdouts for open vocabulary experiments.
}
}
\end{table}

The main challenge with evaluating tag prediction is the sparse and noisy nature of user-generated ground truth.
For instance, a certain movie tag having zero votes may still be relevant in reality. On the other hand, some entities may have votes for contradictory tags (e.g., both ``funny'' and ``not funny''). 
The original Tag Genome baseline \citep{taggenome} mitigated this by collecting an additional dataset of unbiased movie-tag relevance scores. 
Alas, that data has not been released.
Instead, we propose two complementary metrics that cast tag prediction either as binary classification or as a ranking problem.

For classification, we binarize labels as follows. Let $\#(m, t)$ be the number of users who assigned a tag $t$ to a movie $m$. Then its binary counterpart $l(m ,t)$ is set to 1 iff $\#(m, t) > T$, a threshold\footnote{We use $T = 2$ to filter out noisy tags.}.

For the tag ranking formulation, we make the assumption that true movie-tag relevance is correlated with the number of movie-tag votes, and define our movie-tag relevance score as $r(m ,t) = \#(m, t)$. 

Equipped with this score, we use Precision@k and NDCG metrics~\cite{ndcg} to measure performance.

\paragraph{Tag prediction baselines} include
\vspace{-5pt}
\begin{description} 
    \item \toptags --- a fixed ordering of tags.
    \vspace{-5pt}
    \item \tfidf scores for movie-tag pairs, based on tag frequencies in movie reviews.
    \vspace{-5pt}
    \item \bagofbert, as defined in Sec.~\ref{eval:baselines}, is fine-tuned to estimate sentence-to-tag relevance directly\footnote{We found it is best to encode a review sentence using \bert's {\em [CLS]} output, while tags are encoded by averaging individual tokens' output vectors.}. This setup is applicable to both open and closed vocabulary scenarios. During inference, a movie-tag prediction is obtained by averaging over sentence-wise predictions for the movie's reviews.
    \vspace{-5pt}
    \item \textit{\taggenome}---the original baseline from MovieLens team~\citep{taggenome}. The comparison is not entirely apt as that model was trained on additional movie-tag relevance data and user ratings, albeit with a smaller corpus of unsupervised reviews. Also, \taggenome was trained on all of MovieLens (no holdouts). 
    \vspace{-5pt}
    \item Humans---to simulate human performance, apply cross-validation to ground truth user votes, treating one of the folds as a quasi-model.
\end{description}
All models were evaluated on the same holdout sets, with averaging.

\paragraph{Closed Vocabulary Tag Prediction}
In this scenario, evaluation is done on a holdout set of movies (with a smaller development set used for hyperparameter tuning; see Table~\ref{table:movielens_eval_data} for details).

% !TEX root = main.tex

\begin{table}
\centering
\begin{tabular}{lrr}
    \textbf{Model} & \textbf{MAP} & \textbf{AUC} \\
    \hline
    \toptags & 6.2 & 0.80\\
    TD-IDF & 32.3 & 0.86\\    
    \bagofbert & 39.3 & 0.91 \\
    \taggenome & 43.9 & \textbf{0.98} \\
    \hline
    \bertysimple & \textbf{44.7} & \textbf{0.98} \\
    \bertydual & 38.6 & 0.96 \\
    \bertyhybrid & 44.1 & \textbf{0.98} \\
    \hline
    Human & 76.6 & 0.99 \\  
\end{tabular}
\caption{\small Mean Average Precision and ROC-AUC results on the closed-vocabulary tag prediction task. \taggenome is the original baseline from MovieLens creators~\citep{taggenome}, trained on multiple additional features and considered SOTA. Despite using fewer features, \berty matches \taggenome performance on AUC and outperforms it on precision (MAP).
}
\label{table:tap-prediction:closed-vocab}
\end{table}

% !TEX root = main.tex

\setcounter{figure}{2}
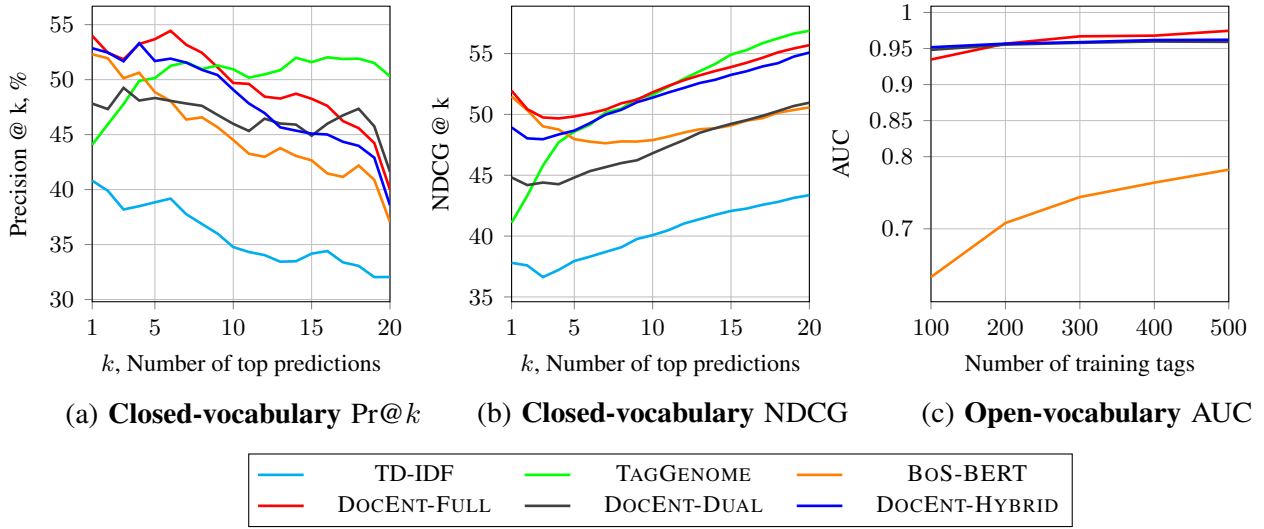
\begin{figure*}
\renewcommand\thesubfigure{(\alph{subfigure})}
\centering
\begin{tikzpicture}
\begin{groupplot}[
    group style={
        group name=ml plots,
        group size=3 by 1,
        ylabels at=edge left,
        horizontal sep=1.6cm
    },
    try min ticks=5,
    footnotesize,
    width=5.5cm,
    height=5.5cm,
    tickpos=left,
    ytick align=outside,
    xtick align=outside,
    enlarge x limits=false,
]
\nextgroupplot[
    xlabel={\small $k$, Number of top predictions},
    ylabel={Precision @ k, \%},
    grid=both,
    xtick={1, 5, 10, 15, 20},
    grid style={line width=.1pt, draw=gray!10},
    major grid style={line width=.2pt,draw=gray!50}    
]
%     \addplot[style={olive,fill=none,mark=none}] coordinates {
% 		(1, 79.57)
% 		(2, 86.26)
% 		(3, 88.69)
% 		(4, 91.37)
% 		(5, 91.62)
% 		(6, 92.82)
% 		(7, 93.53)
% 		(8, 92.75)
% 		(9, 93.39)
% 		(10, 93.73)
% 		(11, 94.75)
% 		(12, 94.96)
% 		(13, 94.56)
% 		(14, 95.30)
% 		(15, 95.03)
% 		(16, 94.39)
% 		(17, 94.21)
% 		(18, 94.04)
% 		(19, 93.99)
% 		(20, 94.29)
%     };
    % \addlegendentry{Human}
	\addplot[style={cyan,fill=none,mark=none}] coordinates {
		(1, 40.81)
		(2, 39.89)
		(3, 38.19)
		(4, 38.49)
		(5, 38.84)
		(6, 39.18)
		(7, 37.76)
		(8, 36.87)
		(9, 35.99)
		(10, 34.77)
		(11, 34.32)
		(12, 34.04)
		(13, 33.44)
		(14, 33.48)
		(15, 34.17)
		(16, 34.41)
		(17, 33.38)
		(18, 33.06)
		(19, 32.04)
		(20, 32.05)
	};
% 	\addlegendentry{tf-idf}		
    \addplot[style={red,fill=none,mark=none}] coordinates {
		(1, 53.99)
		(2, 52.45)
		(3, 51.85)
		(4, 53.24)
		(5, 53.68)
		(6, 54.44)
		(7, 53.17)
		(8, 52.45)
		(9, 51.12)
		(10, 49.70)
		(11, 49.61)
		(12, 48.45)
		(13, 48.28)
		(14, 48.72)
		(15, 48.26)
		(16, 47.62)
		(17, 46.23)
		(18, 45.59)
		(19, 44.23)
		(20, 40.00)
	};
% 	\addlegendentry{\bertysimple}	
    \addplot[style={green,fill=none,mark=none}] coordinates {
		(1, 44.08)
		(2, 45.94)
		(3, 47.77)
		(4, 49.89)
		(5, 50.14)
		(6, 51.25)
		(7, 51.58)
		(8, 50.96)
		(9, 51.29)
		(10, 50.94)
		(11, 50.18)
		(12, 50.48)
		(13, 50.86)
		(14, 51.99)
		(15, 51.59)
		(16, 52.03)
		(17, 51.88)
		(18, 51.91)
		(19, 51.51)
		(20, 50.30)
	};	
% 	\addlegendentry{TagGenome}	
    \addplot[style={orange,fill=none,mark=none}] coordinates {
		(1, 52.31)
		(2, 51.94)
		(3, 50.13)
		(4, 50.63)
		(5, 48.86)
		(6, 48.05)
		(7, 46.37)
		(8, 46.57)
		(9, 45.67)
		(10, 44.51)
		(11, 43.26)
		(12, 42.98)
		(13, 43.77)
		(14, 43.06)
		(15, 42.66)
		(16, 41.47)
		(17, 41.15)
		(18, 42.19)
		(19, 40.88)
		(20, 37.05)
	};		
% 	\addlegendentry{\bagofbert}	
    \addplot[style={darkgray,fill=none,mark=none}] coordinates {
		(1, 47.82)
		(2, 47.32)
		(3, 49.25)
		(4, 48.10)
		(5, 48.33)
		(6, 48.07)
		(7, 47.83)
		(8, 47.62)
		(9, 46.79)
		(10, 45.98)
		(11, 45.33)
		(12, 46.45)
		(13, 46.01)
		(14, 45.91)
		(15, 44.90)
		(16, 45.99)
		(17, 46.75)
		(18, 47.35)
		(19, 45.75)
		(20, 41.62)
	};
% 	\addlegendentry{\bertydual}	
    \addplot[style={blue,fill=none,mark=none}] coordinates {
		(1, 52.86)
		(2, 52.46)
		(3, 51.67)
		(4, 53.31)
		(5, 51.69)
		(6, 51.91)
		(7, 51.55)
		(8, 50.89)
		(9, 50.42)
		(10, 49.08)
		(11, 47.83)
		(12, 46.96)
		(13, 45.66)
		(14, 45.36)
		(15, 45.09)
		(16, 45.01)
		(17, 44.35)
		(18, 43.98)
		(19, 42.90)
		(20, 38.57)
	};		
\nextgroupplot[
    xlabel={\small $k$, Number of top predictions},
    ylabel={NDCG @ k},
    grid=both,
    xtick={1, 5, 10, 15, 20},
    ytick={35, 40, 45, 50, 55, 60},
    legend style = {
        column sep = 10pt,
        legend columns = 3,
        legend to name = grouplegend},
    legend to name=central
]
    \addplot[style={cyan,fill=none,mark=none}] coordinates {
		(1, 37.80)
		(2, 37.59)
		(3, 36.63)
		(4, 37.22)
		(5, 37.95)
		(6, 38.31)
		(7, 38.70)
		(8, 39.08)
		(9, 39.76)
		(10, 40.08)
		(11, 40.48)
		(12, 41.03)
		(13, 41.39)
		(14, 41.75)
		(15, 42.07)
		(16, 42.26)
		(17, 42.57)
		(18, 42.81)
		(19, 43.14)
		(20, 43.36)
	};
	\addlegendentry{TD-IDF}    
    \addplot[style={green,fill=none,mark=none}] coordinates {
		(1, 41.10)
		(2, 43.32)
		(3, 45.79)
		(4, 47.70)
		(5, 48.58)
		(6, 49.15)
		(7, 50.16)
		(8, 50.48)
		(9, 51.27)
		(10, 51.61)
		(11, 52.28)
		(12, 52.93)
		(13, 53.55)
		(14, 54.15)
		(15, 54.92)
		(16, 55.29)
		(17, 55.85)
		(18, 56.25)
		(19, 56.64)
		(20, 56.87)
	};
	\addlegendentry{\taggenome}
    \addplot[style={orange,fill=none,mark=none}] coordinates {
		(1, 51.47)
		(2, 50.35)
		(3, 49.02)
		(4, 48.76)
		(5, 47.97)
		(6, 47.76)
		(7, 47.63)
		(8, 47.78)
		(9, 47.77)
		(10, 47.89)
		(11, 48.18)
		(12, 48.51)
		(13, 48.78)
		(14, 48.86)
		(15, 49.09)
		(16, 49.49)
		(17, 49.69)
		(18, 50.14)
		(19, 50.36)
		(20, 50.58)
	};		
	\addlegendentry{\bagofbert}		
% 	\addplot[style={olive,fill=none,mark=none}] coordinates {
% 		(1, 80.26)
% 		(2, 82.27)
% 		(3, 83.03)
% 		(4, 83.32)
% 		(5, 83.28)
% 		(6, 83.43)
% 		(7, 83.50)
% 		(8, 83.49)
% 		(9, 83.47)
% 		(10, 83.57)
% 		(11, 83.67)
% 		(12, 83.80)
% 		(13, 83.98)
% 		(14, 84.07)
% 		(15, 84.16)
% 		(16, 84.21)
% 		(17, 84.25)
% 		(18, 84.27)
% 		(19, 84.30)
% 		(20, 84.31)
% 	};
% 	\addlegendentry{Humans}
    \addplot[style={red,fill=none,mark=none}] coordinates {
		(1, 51.95)
		(2, 50.43)
		(3, 49.75)
		(4, 49.67)
		(5, 49.83)
		(6, 50.08)
		(7, 50.40)
		(8, 50.91)
		(9, 51.21)
		(10, 51.82)
		(11, 52.34)
		(12, 52.81)
		(13, 53.21)
		(14, 53.57)
		(15, 53.89)
		(16, 54.25)
		(17, 54.65)
		(18, 55.10)
		(19, 55.42)
		(20, 55.69)
	};
	\addlegendentry{\bertysimple}	
    \addplot[style={darkgray,fill=none,mark=none}] coordinates {
		(1, 44.82)
		(2, 44.19)
		(3, 44.40)
		(4, 44.26)
		(5, 44.81)
		(6, 45.34)
		(7, 45.67)
		(8, 46.00)
		(9, 46.23)
		(10, 46.81)
		(11, 47.36)
		(12, 47.89)
		(13, 48.49)
		(14, 48.87)
		(15, 49.22)
		(16, 49.53)
		(17, 49.88)
		(18, 50.28)
		(19, 50.69)
		(20, 50.96)
	};
	\addlegendentry{\bertydual}	
    \addplot[style={blue,fill=none,mark=none}] coordinates {
		(1, 48.93)
		(2, 48.03)
		(3, 47.96)
		(4, 48.35)
		(5, 48.67)
		(6, 49.28)
		(7, 49.98)
		(8, 50.38)
		(9, 51.00)
		(10, 51.38)
		(11, 51.80)
		(12, 52.18)
		(13, 52.58)
		(14, 52.85)
		(15, 53.25)
		(16, 53.55)
		(17, 53.94)
		(18, 54.21)
		(19, 54.75)
		(20, 55.08)
	};		
	\addlegendentry{\bertyhybrid}	
\nextgroupplot[
    xlabel={\small Number of training tags},
    ylabel={AUC},
    grid=both,
    grid style={line width=.1pt, draw=gray!10},
    major grid style={line width=.2pt,draw=gray!50},
    xtick={100,200,300,400,500},
    ytick={0.7, 0.8, 0.85, 0.9, 0.95, 1.0},
]
    \addplot[style={orange,fill=none,mark=none}] coordinates {
        (100, 0.633)
        (200, 0.708)
        (300, 0.744)
        (400, 0.764)
        (500, 0.782)
        % (528, 0.786)
	};
% 	\addlegendentry{\bagofbert}	
    \addplot[style={red,fill=none,mark=none}] coordinates {
        (100, 0.934715)
        (200, 0.956744)
        (300, 0.966907)
        (400, 0.967873)
        (500, 0.974555)
	};	
% 	\addlegendentry{\bertysimple}
    \addplot[style={darkgray,fill=none,mark=none}] coordinates {
        (100, 0.948123)
        (200, 0.955378)
        (300, 0.957920)
        (400, 0.959700)
        (500, 0.959229)
    };	
% 	\addlegendentry{\bertydual}
    \addplot[style={blue,fill=none,mark=none}] coordinates {
        (100, 0.951774)
        (200, 0.956754)
        (300, 0.958808)
        (400, 0.961774)
        (500, 0.962151)
	};	
% 	\addlegendentry{\bertyhybrid}	
\end{groupplot}
\node at ([yshift=-2.5cm]ml plots c2r1.south){\pgfplotslegendfromname{grouplegend}}; 
\node[text width=5cm,align=center,anchor=north] at ([xshift=0.1cm,yshift=-1.2cm]ml plots c1r1.south) {
{(a) \textbf{Closed-vocabulary} Pr@$k$
\label{fig:tag-prediction:closed-vocab:precision}}};
\node[text width=5cm,align=center,anchor=north] at ([xshift=0.1cm,yshift=-1.2cm]ml plots c2r1.south) {{(b) \textbf{Closed-vocabulary} NDCG
\label{fig:tag-prediction:closed-vocab:ndcg}}};
\node[text width=5cm,align=center,anchor=north] at ([xshift=0.1cm,yshift=-1.2cm]ml plots c3r1.south) {{(c) \textbf{Open-vocabulary} AUC\label{fig:tag-prediction:open-vocab}}};
\end{tikzpicture}
\vspace{-0.6cm}
\caption*{
    \label{fig:tag-prediction}
    \hspace{-2pt}Figure 2: \small Performance on tag prediction tasks. Left and center: Precision and NDCG @$k$, with a closed vocabulary. \bertysimple dominates the strong \taggenome baseline for smaller values of $k$, a concentration of gains typical for binary classification models. For perspective, human Precision@$k$ ranges 80-95\% for this task. Right: AUC for open vocabulary experiments, with models trained using a variable fraction of the tag vocabulary. \berty approaches close-vocabulary AUC after training with only 10-50\% of the vocabulary (showing all baselines that were available to us in this setting).
}
\end{figure*}

Results for ranking (MAP) and binary classification (AUC) metrics are shown in Table~\ref{table:tap-prediction:closed-vocab}. Collectively, \berty models outperform the strong \taggenome baseline on tag ranking (see also Fig.~2~(a)~and~(b)) and match (or slightly outperform) it in binary classification. It is a strong result, considering that \berty had no access to additional features used by \taggenome and employed no feature engineering.
Of the three models, \bertydual scores the lowest on all metrics, likely due to not optimizing for $P(Text \mid Entity)$ in pre-training. 
Finally, note that all models still score way below humans on the (harder) tag ranking task, indicating considerable headroom.

\paragraph{Open Vocabulary Tag Prediction}
This task is evaluated by withholding parts of the tag vocabulary so that those tags are never seen in training (consult Table~\ref{table:movielens_eval_data} for details).
Fig.~2~(c) shows our models' performance on the binary classification task base on the fraction of the vocabulary seen by a model in fine-tuning. The graph shows that training with only 100 of the 1124 tags results in reasonable performance. Of our three models, \bertysimple starts below the others but adapts the fastest, reaching a near-closed vocabulary performance with less than 50\% of the full tag vocabulary.

\subsubsection{Reddit Movie Suggestions} 
\label{eval:reddit}
% !TEX root = main.tex

{\small 
\begin{table} %[t!]
\centering
\begin{tabular}{l|r|rr}
\multirow{2}{*}{\textbf{Model}} & \multirow{2}{*}{\textbf{MRR}} & \multicolumn{2}{c}{\textbf{Recall}, \%} \\
& & @50 & @100 \\
\hline
Lucene (\tfidf) & 0.14 & 15.3 & 20.7 \\
\bagofglove & 0.04 & 4.1 & 6.6 \\
\bagofbertave & 0.08 & 9.6 & 14.2 \\
\bagofsbert & 0.07 & 7.6 & 11.7 \\
\hline
\bertysimple & 0.22 & 21.3 & 28.4 \\
% EGO
\bertydual & 0.27 & 28.0 & 36.3 \\
% EGOMASK
\bertyhybrid & \textbf{0.31} & \textbf{31.9} & \textbf{40.9} \\
\end{tabular}
\caption{\label{table:reddit_results} \small Zero-shot results for \berty models vs several baselines on Reddit Movie Suggestions. MRR stands for Mean Reciprocal Rank.
}
\end{table}
}

\paragraph{Movie suggestion baselines} Since this is a search task, we compare our models to an Apache Lucene\footnote{\url{https://lucene.apache.org/}} baseline, arguably the world's most widely used open-source search engine. For completeness, we also compare to \bagofbertave\footnote{In absence of a fine-tuned \textit{[CLS]} output, this version of \bagofbert encodes sentences by averaging their individual tokens' output vectors.}, \bagofglove and \bagofsbert, neural baselines defined in Sec.~\ref{eval:baselines}, whose query-movie relevance score is given by the maximum cosine similarity among the movie's review sentences\footnote{In this case, we found that aggregating sentence-wise predictions with $L^\infty$ norm is superior to averaging.}.

Table~\ref{table:reddit_results} shows the Mean Reciprocal Rank (MRR) as well as recall, metrics that suit the noisy ground truth (for completeness, see also the qualitative results in Table~\ref{table:reddit_qual}). \berty models outperform the Lucene baseline on all metrics, with \bertyhybrid leading by a large margin. Compared to \bertydual, its strong performance is not surprising since \bertyhybrid optimizes both $P(Entity \mid Text)$ and $P(Text \mid Entity)$---a combination of tasks that helps avoid overfitting.

Also expected is the relatively weak performance of $\bertysimple$. As discussed in Sec.~\ref{method}, its high-capacity entity representations are prone to overfitting when the number of entities is relatively small. Still, this shortcoming can be remedied by fine-tuning, as evidenced by this model's superior results on tag prediction in Sec.~\ref{eval:movielens}. These results suggest that \bertysimple may be a good choice in semi-supervised scenarios.

% !TEX root = main.tex
\section{Related Work}
\label{related_work}

Much of the prior art in text-based entity understanding is motivated by the {\em Entity Linking} (EL) problem: predict a unique entity from its mention in text, assuming a single right answer.
By contrast, tasks like entity retrieval and tag prediction imply multiple valid matches and emphasize understanding entities through the prism of their attributes, expressed in natural language.
Still, recent EL works propose dual encoder approaches similar to ours~\citep{DBLP:journals/tacl/YamadaSTT17,relic,cheng-roth-2013-relational,DBLP:conf/ijcai/SunLTYJW15,yamada-etal-2016-joint,chang2020pretraining, DBLP:conf/naacl/KobayashiTOI16, DBLP:conf/acl/HeLLZZW13, DBLP:conf/emnlp/GuptaSR17}, with \citet{relic} already discussed in Section~\ref{sec:bertydual}.
Dual encoders have also been explored in zero-shot scenarios ~\citep{gillick2019learning, DBLP:conf/acl/LogeswaranCLTDL19,wu2019zeroshot,DBLP:conf/emnlp/GuptaSR17}, with entity embeddings computed dynamically based on metadata such as dictionary definitions, entity name and/or category.
Others incorporate entity representations directly in the transformer by retrieving from an external memory~\cite{fvry2020entities,peters2019knowledge}.
While clearly useful for EL, e.g., in sentences with multiple entity mentions, the benefits to our applications are unclear.
Finally, there is ERNIE~\cite{DBLP:journals/corr/abs-1904-09223} -- a language model trained with awareness of entity mentions. 
Alas, the lack of explicit entity representation limits its use in our tasks.

% !TEX root = main.tex
\section{Conclusion \& Future Work}
\label{conclusion}

This paper proposes a family of models to learn self-supervised entity representations from large document collections. We motivate these dedicated representations by contrasting them with naive text-as-a-proxy approaches, with clear gains on entity-centric tasks such as natural language search and movie tag prediction. We then show that achieving superior performance requires optimizing both $P(Entity \mid Text)$ and $P(Text \mid Entity)$---in contrast to the baseline \relic model (and similar prior dual encoders) having only a single objective. To that end, we propose two novel models and study them in zero-shot, few-shot and supervised settings. We match or outperform competitive baselines, where available, with little or no fine-tuning.

\paragraph{Future Work}
As shown qualitatively in Sec.~\ref{sec:tasks:reddit-zero-shot}, \berty has the potential for being a hybrid approach to bridge entity retrieval and recommendation, an application worth exploring in depth (e.g., on the MovieLens Recommendation task which can be readily integrated with \berty thanks to {\em Reviews2Movielens}). A larger entity retrieval study with heterogeneous entity types is another useful direction. Lastly, extending \berty to additional entity understanding tasks such as QA and summarization is yet another promising avenue.

\section*{Acknowlegements}
We appreciate the feedback from the reviewers. This work is partially supported by NSF Awards IIS-1513966/ 1632803/1833137, CCF-1139148, DARPA Awards\#: FA8750-18-2-0117, FA8750-19-1-0504,  DARPA-D3M - Award UCB-00009528, Google Research Awards, gifts from Facebook and Netflix, and ARO\# W911NF-12-1-0241 and W911NF-15-1-0484.

%\bibliography{main_bib}
%\bibliographystyle{acl_natbib}

\end{document}